%% file: main.tex
\documentclass[conference]{IEEEtran}

\IEEEoverridecommandlockouts

\usepackage{cite}
\usepackage{amsmath,amssymb,amsfonts}
\usepackage{algorithmic}
\usepackage{graphicx}
\usepackage{textcomp}
\usepackage{tikz}
\usepackage{xcolor}
\def\BibTeX{{\rm B\kern-.05em{\sc i\kern-.025em b}\kern-.08em
    T\kern-.1667em\lower.7ex\hbox{E}\kern-.125emX}}

\newcommand\copyrighttext{%
  \footnotesize \textcopyright This work has been submitted to the IEEE for possible publication. Copyright may be transferred without notice, after which this version may no longer be accessible.}
\newcommand\copyrightnotice{%
\begin{tikzpicture}[remember picture,overlay]
\node[anchor=south,yshift=10pt] at (current page.south) {\fbox{\parbox{\dimexpr\textwidth-\fboxsep-\fboxrule\relax}{\copyrighttext}}};
\end{tikzpicture}%
}

\begin{document}

\title{Understanding Robot Minds: Leveraging Machine Teaching for Transparent Human-Robot Collaboration Across Diverse Groups}

\author{\IEEEauthorblockN{ Suresh Kumaar Jayaraman}
\IEEEauthorblockA{\textit{Robotics Institute} \\
\textit{Carnegie Mellon University}\\
Pittsburgh, USA \\
}
\and
\IEEEauthorblockN{Reid Simmons}
\IEEEauthorblockA{\textit{Robotics Institute} \\
\textit{Carnegie Mellon University}\\
Pittsburgh, USA \\
}
\and
\IEEEauthorblockN{Aaron Steinfeld}
\IEEEauthorblockA{\textit{Robotics Institute} \\
\textit{Carnegie Mellon University}\\
Pittsburgh, USA \\
}
\and
\IEEEauthorblockN{Henny Admoni}
\IEEEauthorblockA{\textit{Robotics Institute} \\
\textit{Carnegie Mellon University}\\
Pittsburgh, USA \\
}
\thanks{This work was supported by the Office of Naval Research award N00014-181-2503.}  
}
\maketitle
\copyrightnotice

\begin{abstract}
In this work, we aim to improve transparency and efficacy in human-robot collaboration by developing machine teaching algorithms suitable for groups with varied learning capabilities. While previous approaches focused on tailored approaches for teaching individuals, our method teaches teams with various compositions of diverse learners using team belief representations, to address personalization challenges within groups. We investigate various group teaching strategies, such as focusing on individual beliefs or the group's collective beliefs, and assess their impact on learning robot policies for different team compositions. Our findings reveal that team belief strategies yield less variation in learning duration and better accommodate diverse teams compared to individual belief strategies, suggesting their suitability in mixed-proficiency settings with limited resources. Conversely, individual belief strategies provide a more uniform knowledge level, particularly effective for homogeneously inexperienced groups. Our study indicates that the teaching strategy's efficacy is significantly influenced by team composition and learner proficiency, highlighting the importance of real-time assessment of learner proficiency and adapting teaching approaches based on learner proficiency for optimal teaching outcomes. 
\end{abstract}


\input{1.introduction}

\input{2.background}

\input{3.methods}

\input{4.simulation_study}

\input{5.results_discussion}

\input{6.conclusion}

\bibliography{references}
\bibliographystyle{IEEEtran}

\end{document}

%% file: 1.introduction.tex
\section{Introduction}\label{sec:intro}

\begin{figure}[t]
    \centering
    \includegraphics[trim={9.5cm 1.8cm 8cm 2.2cm},clip, width=0.45\textwidth]{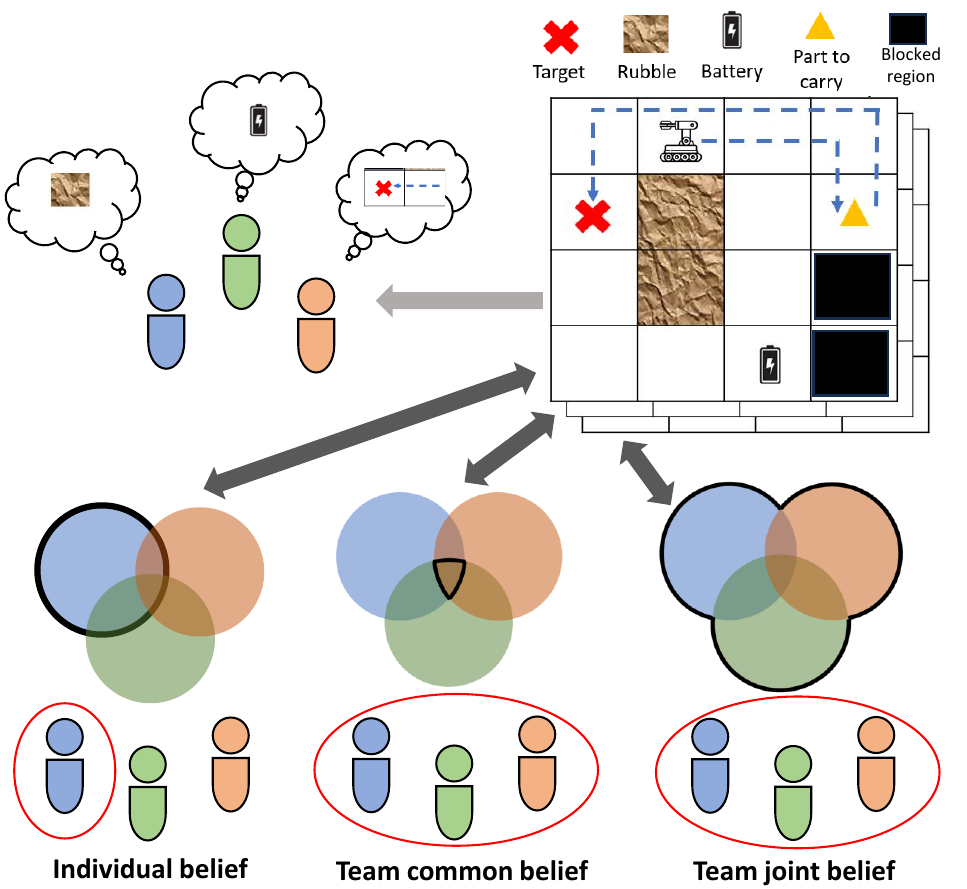}
    \caption{The figure illustrates the complexity of group machine teaching, highlighting the disparity in understanding from common examples among diverse group members. Personalizing examples to a group is challenging due to varied individual beliefs and learning abilities. Our approach utilizes estimations of individual and collective team beliefs to tailor demonstrations for effective communication of the robot's policy to the entire group.}
    \label{fig:intro_figure}
\end{figure}

Robots are increasingly becoming an integral part in people's lives, evolving from human assistants to partners. For safe and effective interaction in human-robot collaboration, it is crucial for humans to understand how robots make decisions. Explainable decision-making aims to clarify the underlying decision-making process of the robot to human collaborators.

Explanations are found to be more useful when they are personalized to the individual \cite{schneider2019personalized, liao2020questioning, silva2024towards}. However, this requires the robots to track the individual's knowledge or understanding of the system and the surroundings to develop explanations that are better suited for them. In prior work, the human is frequently modeled as an inverse reinforcement learner \cite{jara2019theory} and learns a robot policy from demonstrations of its behavior \cite{huang2019enabling}. While traditional machine teaching \cite{zhu2015machine} aims to generate informative demonstrations for individual human learners, real-world scenarios often involve the robot working with diverse groups, introducing the need for teaching methodologies that accommodate groups with varying learning abilities, and experiences. This work investigates approaches for robots to effectively communicate their decision-making to groups of human learners through demonstrations.

Teaching a group as a whole instead of individually is preferable, especially with limited time and resources. Consider, for instance, an ad hoc emergency response team tasked with building shelters after an earthquake. They are given a robot that can bring requested items. The robot has limited maneuverability over rubble, a limited range, and may prefer to recharge when possible. These capabilities and preferences (i.e. its decision-making) must be taught to the team quickly because of the time-sensitive situation. A challenge in group teaching is accommodating individuals with varied learning abilities through a common set of demonstrations. Prior work has shown that it is possible to teach a heterogeneous group of learners using common examples \cite{zhu2017no}, albeit for simple concepts. While groups can also learn from each other through communication and information sharing, we focus only on learning from common examples. Also, group heterogeneity could imply variations in prior knowledge, but we assume similar prior knowledge, focusing solely on differences in the learning ability among the group.

In a classroom, a teacher could personalize the lessons to the class based on various objectives --- focus on naive learners who need more support, or on proficient learners, or consider the class as a whole using class average or similar measures --- and adapt their teaching accordingly. Yeo et al.  \cite{yeo2019iterative} explored categorizing learners based on learning rates and provided personalized teaching to each category. Melo and Lopes \cite{melo2021teaching}, on the other hand, generated personalized demonstrations for each learner, but at a high teaching cost. An active teacher that personalizes and adapts to the learner can improve learning \cite{kamalaruban2019interactive, lee2023closed}. But a challenge in groups is identifying to whom the personalization should cater. 


Drawing inspiration from pedagogical literature on teaching classrooms \cite{carpenter2006effective}, our key insight is that \emph{machine teaching can be tailored to a group of learners by considering the group as a whole and generating demonstrations based on the aggregation of the group's understanding.} In this work, we developed team belief models that facilitate group teaching focusing on the entire team. We utilized a closed-loop teaching framework that effectively incorporates feedback from the robot teacher to aid human learners. We adapted a human belief model to generate simulated human learners of varying learning abilities. We conducted a simulation study to explore how different group teaching strategies affect the group's learning and how team composition of learners with varying learning abilities --- naive and proficient --- moderate group learning. Our findings suggests that teaching methods designed for individual beliefs weren't much affected by how knowledgeable team members were. However, these methods did affect how long team members interacted, depending on team composition. On the other hand, teaching methods that focused on team beliefs helped increase knowledge, especially in groups with more proficient learners. 

%% file: 2.background.tex
\section{Background}\label{sec:background}

\textbf{Machine teaching for policies:} 
We model the environment as a Markov Decision Process (MDP), given by the tuple $\langle \mathcal{S}, \mathcal{A}, T, R, \gamma, \mathcal{S_0} \rangle$, representing the state and action spaces, transition function, reward function, discount factor, and initial state distribution respectively. An optimal trajectory $\xi^*$ is a sequence of $(s_i, a, s_i')$ tuples obtained by following the robot's optimal policy $\pi^*$. Similar to prior work \cite{abbeel2004apprenticeship}, $R = \mathbf{w^*}^\top \phi(s, a, s')$ is represented as a weighted linear combination of reward features. We define a group of MDPs that share $R, \mathcal{A},$ and $\gamma$ but differ in $T_i,\,  \mathcal{S}_i,$ and $S^0_i$, as a domain. Sharing the same $R$ ensures that all demonstrations within the domain support inference over a common $\mathbf{w^*}$. We use the MDP formulation to model an \textit{item delivery} task where a robot is tasked with delivering an item in an environment that has rubble, blockages, and a battery recharge station (see Fig.\ref{fig:intro_figure}). 

We adapt the machine teaching framework for policies \cite{lee2022reasoning} to select a set of demonstrations $\mathcal{D}$ of size $n$ that maximizes the similarity $\rho$ between optimal policy $\pi^*$ and the policy $\hat{\pi}$ recovered using a computational model $\mathcal{M}$ (e.g., IRL) on $\mathcal{D}$, $
\arg \max_{\mathcal{D} \subset \Xi} \rho(\hat{\pi}(\mathcal{D}, \mathcal{M}), \pi^*) \;\; \mathrm{s.t.} \;\; |\mathcal{D}| = n $, where $\Xi$ is the set of all demonstrations of $\pi^*$ in a domain. Once $\mathbf{w}^*$ is approximated through IRL, this approach assumes that the learner can deduce $\pi^*$ by planning on the underlying MDP. Thus, the objective reduces to selecting demonstrations that are informative in conveying $\mathbf{w}^*$, which can be measured using behavior equivalence classes. 

\textbf{Behavior equivalence class:} \textit{The behavioral equivalence class (BEC)} of a policy $\pi$ is the set of reward functions under which $\pi$ is optimal. For a reward function that is a weighted linear combination of features, the BEC of a demonstration $\xi$ of $\pi$ is the intersection of half-spaces \cite{brown2019machine} formed by the exact IRL equation
\begin{equation}
\small
\textrm{BEC}(\xi|\pi) := \mathbf{w}^{\top}\left(\mu_{\pi}^{(s, a)}-\mu_{\pi}^{(s, b)}\right) \geq 0, \forall(s, a) \in \xi, b \in \mathcal{A}.
    \label{eq:BEC_demo}
\end{equation} 
where $\mu_{\pi}^{(s, a)}=\mathbb{E} \left[ \sum_{t=0}^{\infty} \gamma^{t} \phi\left(s_{t}\right) \mid \pi, s_{0}=s, a_{0}=a \right]$ is the vector of reward feature counts accrued from taking action $a$ in $s$, then following $\pi$ after. Any demonstration can be converted into a set of constraints on $\mathbf{w}$ using (\ref{eq:BEC_demo}), with each constraint being a \textit{knowledge component (KC)} \cite{koedinger2012knowledge} that captures a facet of the reward function (e.g., tradeoffs between the underlying reward features). Consider the item delivery domain, which has binary reward features $\mathbf{\phi} =$ [\emph{traversed rubble}, \emph{battery recharged}, \emph{action taken}]. In practice, we require $||\mathbf{w^*}||_2 = 1$ to bypass both the scale invariance of IRL and the degenerate all-zero reward function. If no prior knowledge is assumed, the potential belief space on reward weights would uniformly span the surface of the $n-1$ sphere (where $n$ is number of domain features) due to the $L^2$ norm constraint on $\mathbf{w^*}$. We instead assume that learners begin with a prior that action weight is negative (e.g. favoring shortest path, see Fig. \ref{fig:pf_update}).

\textbf{Team modeling:} A common way to represent a team characteristic such as team knowledge is by aggregating the knowledge of individuals. Team characteristics are normally represented as average, median, sum, range, minimum, or maximum values of the characteristic of individuals \cite{cooke2000measuring}. More recently, team knowledge is represented using a latent \emph{collective intelligence} parameter that is highly correlated with team process and performance \cite{riedl2021quantifying}. However, operationalizing such a latent parameter is challenging and we thus represent team belief through observable behaviors by aggregating individual beliefs. We focus on two aggregated representations of team belief --- common belief and joint belief. We define \textbf {common team belief} as the belief that all team members have. It can be visualized as the intersection of individual beliefs. We define \textbf{joint team belief} as the knowledge that at least one individual in the team has, visualized as the union of individual beliefs (see Fig. \ref{fig:constraint_space} (b) for visual representations of these).





%% file: 3.methods.tex
\section{Methods}\label{sec:methods}

In this section, we discuss an approach using particle filters (PF) for modeling individual and team beliefs about the robot's decision-making, i.e. its reward. We use these different beliefs to select corresponding demonstrations that are shown to the entire team. \cite{lee2023closed} originally proposed a PF-based approach to model individual human belief that supports iterative Bayesian updates and sampling for generating informative and tailored demonstrations using counterfactual reasoning. We extend this approach to group teaching to model aggregated team beliefs in addition to individual beliefs. We use this model in a closed-loop teaching framework leveraging insights from the education literature and adaptively generating demonstrations based on individual and aggregated team beliefs. In addition, after seeing demonstrations, we provide tests, which collect responses on expected optimal robot trajectory in an unseen enrivonment to evaluate their understanding.

\subsection{Particle filter model of human learner belief}


Humans generally perform approximate inference from demonstrations \cite{huang2019enabling} and thus we model the human learner's belief about the robot's reward weights using a particle filter, where each particle represents a potential belief about the reward weights \cite{lee2023closed} and the particle weight represents the belief probability. The particle filter follows a Bayesian update process that uses constraints corresponding to the expected information gain for demonstrations and actual information gain for tests. This formulation enables iterative updates on learner belief from demonstrations and tests.

\begin{figure}[h]
    \centering
    \includegraphics[trim={0cm 9.3cm 14cm 0cm},clip, width=0.5\textwidth]{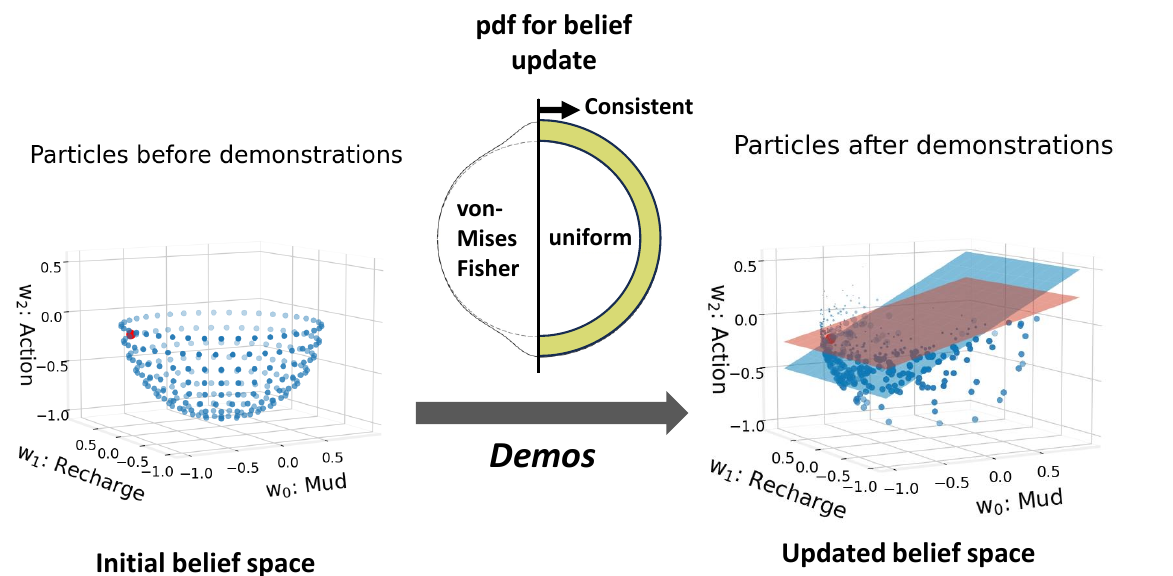}
    \caption{Update process of a learner's belief represented by a particle filter. A cross-section of the custom probability density function (pdf) used to update particle weights is shown. Particles consistent with the demonstrated behavior receive higher weights via a uniform distribution (yellow ring), while those on the inconsistent side are weighted less, decreasing exponentially with distance from the constraint, via a von-Mises Fisher distribution. The updated belief distribution is shown on the right.}
    \label{fig:pf_update}
\end{figure}

From demonstrations, constraints on reward weights can be obtained using Eq. \ref{eq:BEC_demo}, by comparing the optimal demonstration with possible counterfactuals. Similarly, the correct test response can be compared with the incorrect learner response to get these constraints using Eq. \ref{eq:BEC_demo}. Each constraint $c_i$ generated from the demonstrations and test responses is a half-space constraint, meaning, one side is consistent with the demo or test response and the other is not. Each constraint $c_i$ can be converted to a probability distribution $p(x_i | c_i)$ that is used to update the particle weights $w_i$ of particles $x_i$.


We use the custom probability distribution (refer Fig. \ref{fig:pf_update}) proposed in \cite{lee2023closed}, designed such that any particle on the consistent side of the constraint (yellow region in Fig. \ref{fig:pf_update}) is equally valid and could have generated the demonstration (represented by the uniform distribution) and the particles on the inconsistent side are exponentially less likely to have generated the demonstration (represented by the von Mises-Fisher distribution). Fig. \ref{fig:pf_update} shows how the initial learner belief for the delivery robot's reward gets updated using this custom pdf after seeing the demonstrations. The belief updates after demonstrations and feedback are similar to the \emph{predict} stage and the update based on the learner's test response is similar to the \emph{correct} stage of Bayesian estimation. The robot teacher uses the learner belief space to sample possible counterfactuals for generating learner-centric demonstrations. To handle potential sample degeneracy in particle filtering, we add a Gaussian noise $\eta$ when updating the particles \cite{li2014fight}. We also add a small Gaussian noise $\nu$ while updating the particles after test to account for the teacher's estimation noise. For more details on the particle filter model refer \cite{lee2023closed}.


\begin{figure}
    \centering
    \includegraphics[trim={0cm, 8cm, 21.5cm, 0cm}, clip, width=0.35\textwidth]{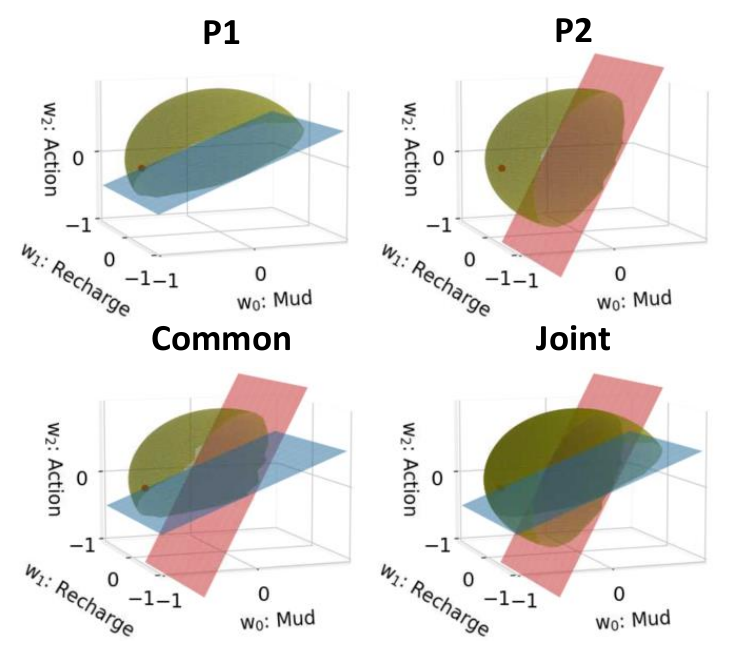}
    \caption{ This figure illustrates an example set of test responses for a team with two individuals, P1 and P2. The test responses are transformed to constraints. The yellow partial spheres show the regions that are consistent with their test response, i.e. agree with the constraint. When their responses are different, the constraints space of their common belief of their tests is their intersection of individual beliefs and that of their joint belief is the union of individual beliefs as depicted. These constraints spaces are used to update the weights of the PF distributions.}
    \label{fig:constraint_space}
\end{figure}

\begin{figure*}[t]
    \centering
    \includegraphics[trim={0.4cm 6.5cm 0.4cm 0cm},clip, width=0.90\textwidth]{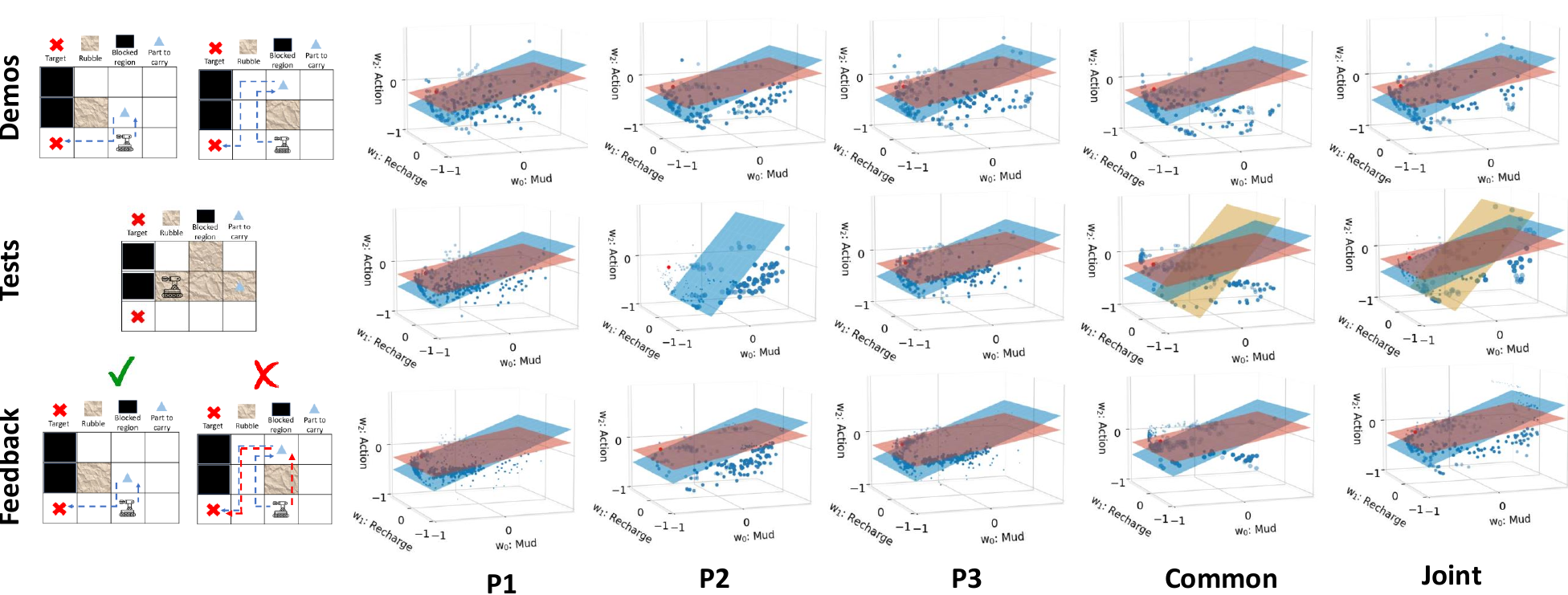}
    \caption{Interactions and corresponding PF belief updates for a team with three people for the first interaction period. The red particle represents the true reward weight. An interaction period consists of one set of demos related to a KC, followed by a set of tests, and then feedback (corrective or confirmatory). After the demos, all individual and team beliefs are updated based on expected information gain from demos. The distributions are similar since all individuals are expected to learn equally. After the demos, they are provided with test(s) to evaluate their understanding and their responses are used directly for updating individual beliefs and aggregated to update team beliefs. In this case, P1 and P3 got the response correctly and P2 got it incorrect. The difference in the constraint spaces and the updated distributions after tests can be observed for the individual and team beliefs. Confirmatory or corrective feedback is given after the tests and they are expected to learn from either feedback. The distributions are updated to reflect this learning from feedback.}
    \label{fig:interaction-period}
\end{figure*}

\subsection{Team belief modeling}

A demonstration's ability to reveal the underlying reward function via IRL is highly dependent on the counterfactuals considered. Thus the learner's belief space from which the counterfactuals are sampled critically influences the demonstrations generated. For groups, we model team belief as aggregations of individual member beliefs \cite{cooke2000measuring, jayaraman2023adaptive}. The main difference in modeling team beliefs is how the particles are updated, specifically how individual constraints are aggregated and used for updating the particle weights.

We envision a group teaching scenario akin to classroom teaching and assume that each team member will have the same kind of interactions, i.e. see the same demonstrations and are provided the same tests. Thus the constraints from the demonstrations will be similar for all individuals. However, let us assume that the team with $m$ members had different responses to a set of $n$ tests, and their constraints are denoted as $\mathcal{C}_1 = \{c^1_1, c^2_1, ... c^n_1\}$, $ \mathcal{C}_2 = \{c^1_2, c^2_2, ..., c^n_2\} $, and $ \mathcal{C}_m = \{c^1_m, c^2_m, ... c^n_m\} $. The update probability for each member is given by, $P_i = \prod_{j=1}^{n} p(x_i^j | c_i^j) $.

We model common team belief by considering the constraints of all members and representing it as $\mathcal{C}_{ck} = \{ c^1_1, c^1_2, ..., c^1_m, c^2_1, c^2_2, ..., c^2_m,  ..., c^n_1, c^n_2, ..., c^n_m \}$. We assume individuals to be independent. Consequently, the particle filter representing common team belief is updated based on the probability of all aggregated constraints across all tests in the set for all the individuals, given by, $P = \prod_{i=1}^{m}\prod_{j=1}^{n} p(x_i^j | c_i^j) $. This aligns with our definition of common belief as the belief that everyone on the team has. 

Joint belief, on the other hand, is modeled by considering the set of constraints for all individuals for each test separately and is represented as a set of disjointed subsets, $\mathcal{C}_{jk} = \{ \{ c^1_1, c^1_2,..., c^1_m \}, \{ c^2_1, c^2_2,..., c^2_m\}, ..., \{ c^n_1, c^n_2,..., c^n_m\} \}$, where each subset represents the constraints of the individuals. Update probabilities are calculated individually for each team member and the particles are updated based on the maximum probability of any of the individuals, given by, $P = \arg \max_{i \in [1, 2, ..., m]} \prod_{j=1}^{n} p(x_i^j | c_i^j) $. This corresponds to our definition of joint belief as the belief that at least one team member has.

Feedback is an effective learning mechanism \cite{fazio2010receiving}. Fig. \ref{fig:interaction-period} shows the effects of the interactions --- demonstrations, tests, and feedback ---  the team has during one interaction period. An interaction period aims to teach a specific KC, for example, the trade-off between mud cost and step cost for the delivery robot. Each interaction set consists of a set of demonstrations, a set of tests, and a set of feedback (corrective or confirmatory feedback based on whether the response was \emph{incorrect} or \emph{correct}). Confirmatory feedback reinforces the learner's knowledge while corrective feedback informs them their learning is incorrect and also provides the correct response.



We can see in Fig. \ref{fig:interaction-period} that for individuals \emph{P1} and \emph{P3} who responded correctly, the belief distribution gets more concentrated within the area of the constraints. This gets further strengthened by the confirmatory feedback they receive. On the other hand, individual \emph{P2} responded incorrectly indicating they may not have learned the KC. Thus their belief distribution moves away from the constraints region of the KC. However, getting corrective feedback brings the distribution closer to the constraints region. The common team belief moves a little further away because of the incorrect response of \emph{P2} but is still mostly close to the correct reward and the constraints region.

\subsection{Teaching curriculum development}
We employ the methods discussed in \cite{lee2022reasoning} to generate demonstrations and tests for teaching the robot's policy. The approach primarily considers likely learner counterfactuals by estimating the learner's belief about robot policy (i.e. reward weights) and sampling $n$ possible counterfactuals from this belief space. For every possible robot demonstration in a domain, and for each reward weight, we simulate what the “human” counterfactual to each demonstration would be if the human had this reward weight in mind and generate the corresponding constraints using Eq. \ref{eq:BEC_demo} and consolidate all these constraints. We select the demonstration from this set of constraints that maximizes knowledge gain before and after seeing the demonstration. We use the consolidated set of constraints to identify test environments that examine the concept taught in the demos (refer \cite{lee2022reasoning} for additional information).

\subsection{Simulated learner model}

Learners have different cognitive capabilities and understand at different levels the same information provided. Furthermore, the teaching process is likely to be an adaptive and varied process, catered to the specific learner. Thus our model of the learner should be able to \emph{encompass a wide variety of learner abilities in different teaching contexts}. We again employ a particle filter-based approach to simulate a learner's learning dynamics and belief updates. Similar to the teacher's model, each particle represents a potential belief about the robot's reward function, but they are the learner's self-belief as opposed to the teacher's estimated learner belief. There are two key differences between the teacher and the learner models --- first, the learner model updates after seeing the demonstration and feedback only and not after tests since we expect that learners will get information only from the demonstrations and  feedback and not from just knowing if they got test responses correct/incorrect \cite{fazio2010receiving} and second, the learner model does not have any estimation Gaussian noise ($\nu$) and only the resampling noise ($\eta$).

Each individual has a different ability to understand the information conveyed through visual demonstrations. We parametrize this ability as  $\beta$ that can vary for each individual. It is operationalized as the probability mass on the uniform (consistent) side of the custom distribution of the underlying constraint (see Fig. \ref{fig:pf_update}). It modifies the information gain in the demonstrations (and feedback) used for particle updates, $ IG_{pf}\,=\, f(\beta, IG_{demo}) $. A higher $\beta$ implies that learners assign more weights to particles on the consistent side of the constraints, indicating certainty over particles that likely resulted in the demonstrations. 


We initialize, $\beta\, =\,\beta_0$, as an individual's ability to learn from demonstrations.  People get better at learning a concept when they get feedback and are repeatedly exposed to the concept \cite{fazio2010receiving}. To incorporate the effects of feedback, we define the feedback dynamics of $\beta$ as,

\vspace{-3mm}

\begin{subequations}
    \begin{equation}
        \beta_t\,=\, \beta_{t-1} + \Delta\,\beta_{t-1}, \,\, \text{with,}
    \end{equation}
    \begin{equation}
        \Delta\,\beta_{t-1} = 
        \begin{cases}
          \delta\,\beta_c & \text{if test at $t-1$ is \emph{correct}}\\
          \delta\,\beta_i & \text{if test at time $t-1$ is \emph{incorrect}}\\
        \end{cases}   
        \label{eq:beta_dynamics}
    \end{equation}
\end{subequations}

where $\beta_c$ is change in $\beta$ due to confirmatory feedback, when the learner's test response is correct and $\beta_i$ is due to corrective feedback, when the learner's test response is incorrect. Corrective and confirmatory feedback have different effects on learning \cite{fazio2010receiving}. Fazio et al. \cite{fazio2010receiving} found that feedback on incorrect responses led to more learning than feedback on correct responses, particularly in more difficult tasks. Thus we define $\beta_i \,>\,  \beta_c$. $\beta_t$ resets to $\beta_0$, for each new concept as we assume the concepts to be independent. Thus improvement in $\beta$ due to feedback is contained within the specific concept or KC.




%% file: 4.simulation_study.tex
\section{Simulation Study}\label{sec:sim_study}

We ran a simulation study to evaluate the effects of different teaching strategies on group learning. The strategies differed in the belief space they utilized for sampling possible counterfactuals to generate informative demonstrations. We used N=8 counterfactuals in this study. 
We also examined the effects of the various team compositions of diverse group members on group learning.

\subsection{Metrics}
We measure the teaching-learning performance using two measures --- (1) the number of interaction periods ($N_i$) taken to learn the policy, and (2) the average team knowledge level at the end of learning. For each individual, their knowledge level is defined as the proportion of belief space that lies within the BEC region of the robot's policy at the end of the learning session. It is given by $p_{BEC} = \sum_{j=1}^{m} p_i,\,  \forall\, i \in \epsilon_{BEC}$, where $j$ is the individual, $\epsilon_{BEC}$ is the BEC region and $i$ indicates an individual particle and $p_i$ its normalized weight. The team knowledge level, $\overline{p_{BEC}}$ is the average knowledge level of all individuals.


\subsection{Study conditions}
The primary condition that we examined was the \emph{teaching strategy} employed to generate the demonstrations and tests. We considered four strategies that samples counterfactuals from four different belief spaces for each interaction period --- \emph{individual low}, the belief space is of the individual with the lowest knowledge about the robot's reward, \emph{individual high}, the belief space is of the individual with the highest knowledge about the robot's reward, \emph{common}, the belief space is the common team belief, and \emph{joint}, where the belief space is the joint team belief. The individual, common and joint belief spaces are visualized in Fig. \ref{fig:constraint_space}. The bigger the belief space, the more diverse the sampled counterfactuals would be. The individuals with lowest or highest knowledge are identified at the end of each interaction period and their corresponding belief spaces are used for the next interaction period corresponding to the strategy employed. We compared these different strategies with a baseline strategy of separately teaching each individual sequentially.

We also examined the influence of \emph{team composition} on the group's learning. Particularly, we considered two categories of learners, \emph{novice} and \emph{proficient}. Novice learners are considered to be beginners and have a low ability to learn from demonstrations, given by a low $\beta_0$. On the other hand, proficient learners have a higher $\beta_0$. For the simulation study, we estimate the distribution of the learner parameters for both the types of learners (see Table \ref{tab:sim_params}). We considered four team compositions, \emph{all novice}, \emph{majority novice}, \emph{majority proficient}, and \emph{all proficient}.

We expect that group teaching strategies based on group belief, especially the ones based on team belief, target the group as a whole and hence would be able to cater to the entire group resulting in faster group teaching. However, individual teaching strategies, specifically ``individual low'' is likely to result in higher knowledge as it focuses on the learners who have the least knowledge. So demonstrations catered for them will also improve the knowledge of others. We also expect that teams with more `Proficient' learners will learn quicker than other teams. We expect that the baseline strategy, which teaches each learner separately would take more interactions, would result in better learning because it personally focuses on each learner. Thus we expect, \emph{H1: Group-belief based strategies to have fewer interactions than individual-belief based strategies.} \emph{H2: Individual low strategy to have the highest knowledge level apart from baseline because of its personalization.} \emph{H3: Teams with more high learners to have both learn faster and high knowledge levels.}

\subsection{Study setup}
We conducted the study for a team with 3 learners. We ran a 5 $\times$ 4 study for the two conditions of \emph{teaching strategy} and \emph{team composition} including the baseline strategy. For each combination, we collected 15 `simulated' teams' data totaling to 300 teams.

\begin{itemize}
    \item \emph{Teaching strategy}: baseline, individual low, individual high, common, and joint
    \item \emph{Team composition}: [N, N, N], [N, N, P], [N, P, P], and [P, P, P], where `N' denotes Novice and `P' denotes Proficient learners.
\end{itemize}

The learning parameters for each type of learner was estimated (see Table \ref{tab:sim_params}) from human learner data collected from a user study discussed in \cite{lee2023closed}. We performed a grid search of the parameters by simulating the teaching interactions from the dataset for all possible grid combinations and choosing the runs that have performance error with $p_{BEC} \, < \, \epsilon$

\begin{table}[h]
    \centering
    \begin{tabular}{c|c|c|c}
       Learner type  & $\beta_0\, (\mu\, , \sigma)$  & $\delta \beta_c \,(\sigma)$ & $\delta \beta_i\, (\sigma)$ \\
       \hline 
       Novice & 0.703 (0.034) & 0.033 & 0.056 \\
       Proficient & 0.809 (0.025) & 0.022 & 0.052
    \end{tabular}
    \caption{Learner parameters estimated from human learner data.}
    \label{tab:sim_params}
\end{table}


The demonstrations are selected based on the KCs and the teaching strategy (which individual/team belief to adapt). The tests are similarly identified to evaluate the constraints conveyed by the demonstrations. For the conditions based on individual beliefs, the individuals with the lowest and highest knowledge are identified at the end of each interaction period, where an interaction period consists of a set of demonstrations, tests, and feedback related to a specific KC. The teaching moves to the next KC only after all individuals in the team learn the current KC. If the team does not learn the KC, the interaction loop continues for the same KC and the demonstrations are generated based on the updated individual or group belief. For aggregated group belief, for example, common belief, it is possible that the responses of teammates are such that there is no common intersecting constraint region. In such cases, we exclude the conflicting individual(s) constraint spaces and consider the plausible intersecting region formed by most team members.

We developed a closed-loop teaching framework to sequentially generate demonstrations, tests, and feedback to teach and evaluate the team's understanding of the robot policy. Utilzing scaffolding techniques discussed in \cite{lee2022reasoning}, we select and sequentially introduce knowledge components (KCs). KCs are broadly defined in the education literature as ``a concept, principle, fact, or skill inferred from performance on a set of related tasks'' \cite{koedinger2012knowledge}. In our case, KCs represent specific characteristics or distinct constraints of the reward features. For example, the KCs could incrementally teach the bounds on the cost of traveling through rubble given the step cost, followed by bounds on the reward for recharging given the step cost, and then trade-offs between rubble and battery. 

After the demonstrations, the group members are given test(s) related to the current KC, to evaluate their understanding. Ideally, human learners will be provided a test environment (see Fig. \ref{fig:interaction-period}) and asked to map the trajectory they believe the robot will take. For our simulation study, we sample a reward weight based on the individual's PF distribution and use that as the response to the test environment. The more proportion of particles that lie within the consistent area of the test environment, the more likely the sampled weight vector gets the test response correct. In case they get the response correct, a confirmatory feedback is provided and if they get the response incorrect, a corrective feedback is provided. The $\beta_t$ of the individual learners are updated according to Eq. \ref{eq:beta_dynamics}.

%% file: 5.results_discussion.tex
\begin{figure*}[t]
    \centering
    \includegraphics[trim={0.5cm 7.6cm 0.3cm 0.2cm},clip, width=0.75\textwidth]{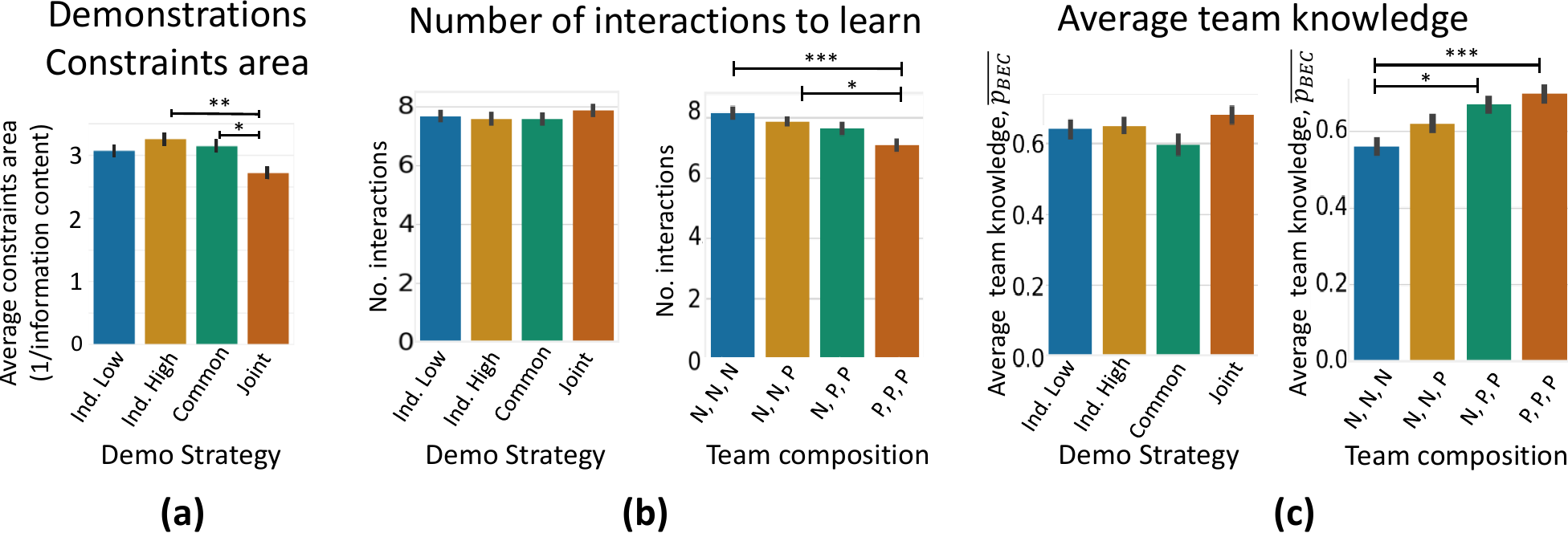}
    \caption{Experimental results on the effects of demonstration strategy and team composition. (a) All group teaching strategies performed better than the baseline strategy of teaching individuals sequentially in terms of number of interactions. No discernable difference due to strategies for number of interactions. Expected differences in teams, teams with more proficient learners learned quicker, observed. (b) Noticeable differences in average team knowledge observed for strategy but differences are not statistically significant. Also observed that teams with more proficient learners had higher knowledge level.}
    \label{fig:results-exp}
\end{figure*}

\section{Results and discussion}\label{sec:results}

\textbf{Manipulation check:} We wanted to ascertain the distinctiveness of the demonstrations and information provided by the various demonstration strategies. We used the surface area formed by the constraints space (yellow region in Fig. \ref{fig:constraint_space}) of the demonstrations at each interaction period for comparison, the lower the area, the higher the information conveyed by the demonstrations. Fig. \ref{fig:results-exp} (a) illustrates that the constraints area for the various demonstration strategies are significantly different ($p<0.01$). Demonstrations based on joint belief result in the most informative demonstrations, whereas those based on the belief of the individual with highest knowledge result in the least informative demonstrations. This is likely because joint belief is a union of individual beliefs, and has a broader distribution resulting in diverse counterfactuals which in turn generate more informative demonstrations.


\textbf{Experimental conditions results:} For the four group teaching strategies, the average number of interactions to learn the reward weights is $ N_g= 7.67 (1.68) $. Unsurprisingly, the baseline strategy of teaching each learner separately takes more interactions, almost twice as much, $N_b= 17.13 (2.32)$. However, contrary to our expectations, the baseline condition had a lower knowledge level, $k_b=0.59 (0.12)$ than the group conditions, $k_g= 0.64 (0.21)$, though the difference is not statistically significant. 


\begin{figure}[h]
    \centering
    \includegraphics[trim={0.4cm 6cm 8cm 0.5cm},clip, width=0.42\textwidth]{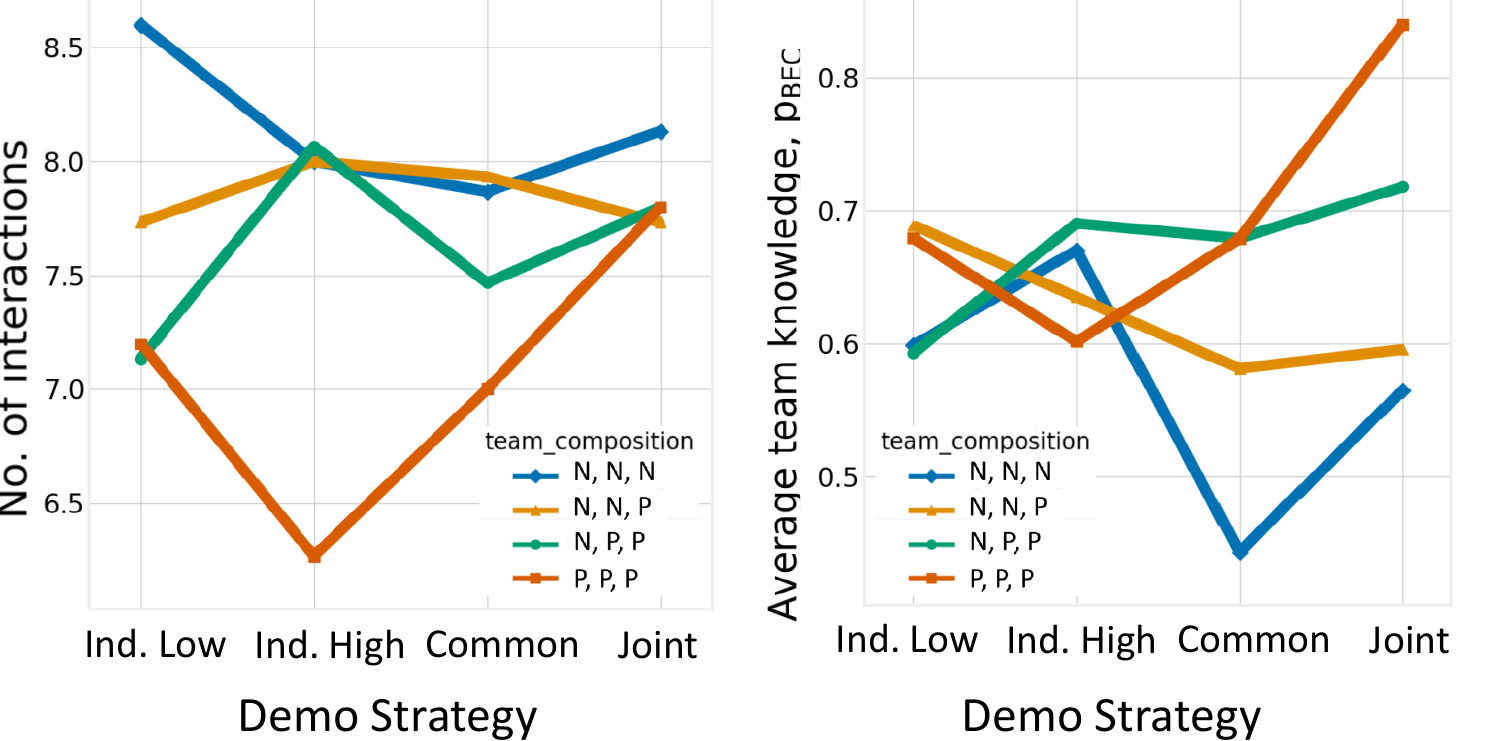}
    \caption{Interaction effects of demo strategy and team composition on number of interactions and average team knowledge. Differences in variance individual and group belief for both the metrics can be observed.}
    \label{fig:results-interaction}
\end{figure}

Since the baseline condition is distinctly different than the other group teaching strategies, we performed two-way ANOVA only on the four group teaching strategies to examine their effects on team learning (see Fig. \ref{fig:results-exp} (b) and (c)). Demonstration strategy did not significantly influence either the number of interactions $(F=0.42,\,p=0.74)$ or team knowledge level $(F=1.73,\,p=0.16)$. Team composition, on the other hand, significantly influences both the number of interactions $(F=4.67,\,p=0.00)$ and team knowledge level $(F=4.64,\,p=0.00)$, as expected. This is not surprising as with more proficient learners in the team, the team is likely to learn faster and better. We also found a significant interaction effect between demonstration strategy and team composition for team knowledge level $(F=2.32.,\,p=0.02)$.



Fig. \ref{fig:results-interaction} shows the interaction trends between demonstration strategy and team composition. While there are no significant interaction effects on the number of interactions, some interesting trends are observed. Individual belief strategies have more variance in learning duration compared to group belief strategies. Group belief strategies, particularly joint belief strategy, is able to accommodate diverse teams and have similar teaching durations. Group belief strategies might therefore be more suitable in situations where the proficiencies of the learners are unclear but still the group has to be taught with limited teaching resources. 

On the other hand, we find significant interaction effects for team knowledge level. The interaction trends, are however, interesting as there is less variance in the team's knowledge level for individual belief and more variation for group belief. This could be because of the targeted nature of the individual belief demonstrations, which adaptively samples counterfactuals for upcoming demonstrations from individual belief spaces based on their current knowledge level. This could result in bringing a uniformity of knowledge in the individual belief conditions, particularly for the `individual low' condition. 

To further understand the significant interaction effect we found for team knowledge level, we perform one-way ANOVA for the strategy condition for each team combination and found that demonstration strategy significantly influenced (after applying Bonferroni correction) team knowledge level for homogeneous teams [N,N,N] ($p < 0.05$) and [P,P,P] ($p < 0.05$). Specifically, group belief strategies work well for teams with all proficient learners whereas individual belief strategies work well for teams with all naive learners. For mixed teams, while we do not observe significant differences, we observe similar trends. For teams with more proficient learners perform better with group strategies while teams with more naive learners perform better with individual strategies.  The robot can choose the appropriate strategy from the start if proficiency information is available apriori but is more robust if it starts with a strategy and then adaptively change the strategy based on real-time information about the learners proficiency. These nuanced results thus highlight the need to observe and estimate learner proficiency in real-time for more effective teaching.

%% file: 6.conclusion.tex
\section{Conclusion}\label{sec:conclusion}

In this study, we aimed to enhance the transparency and efficacy of human-robot collaboration among human groups through explainable robot demonstrations. Our approach involved developing machine teaching algorithms that cater to teams with diverse learning abilities, employing team belief representations aggregated from individual beliefs represented through particle filters. This approach aimed to overcome the personalization challenges present within heterogeneous groups. Our findings revealed that teaching strategies tailored to group or individual beliefs significantly benefit distinctly different groups characterized by varying levels of learner capabilities. Specifically, we observed that the group belief strategy and joint belief, in particular, was advantageous for groups composed mostly of proficient learners. Individual strategies were better suited for groups with mostly naive learners, though they would take more interactions. We gained deeper insights into the dynamics of group learning, thus laying the foundation for adaptively selecting teaching strategies to facilitate collaborative decision-making in real-time scenarios. However, our study has several limitations. In particular, the simulation did not evaluate the teaching algorithms across multiple domains, nor did it involve actual human learners. Our study had isolated simulated learners and does not consider the nuances of interaction within the group. These limitations highlight the simulation-centric nature of our investigation and suggest the need for empirical validation in real-world settings. Moving forward, we plan to extend this simulation study by conducting a human-subjects user study. This future research will involve actual human learners with diverse learning abilities to assess the efficacy of our teaching strategies. Additionally, we aim to explore the applicability of our methods across various domains and settings to ensure generalizability. By addressing the interaction dynamics within groups, we aspire to refine our teaching algorithms further, ensuring that they are not only effective but also adaptable to the needs of different types of learners and group compositions. By continuing to explore and refine machine teaching approaches, we anticipate contributing to the development of robots that can seamlessly integrate into human teams, enhancing both efficiency and understanding in collaborative tasks.
